\documentclass[10pt,twocolumn,letterpaper]{article}

\usepackage{cvpr}
\usepackage{times}
\usepackage{epsfig}
\usepackage{graphicx}
\usepackage{amsmath}
\usepackage{amssymb}
\usepackage{mathtools}
\usepackage{multirow}
\usepackage{authblk}
\usepackage[symbol]{footmisc}
\usepackage{multirow}
\usepackage{ctable} 
\usepackage{subfig}
\usepackage{floatrow}

\def\bal#1\eal{\begin{align}#1\end{align}} %

\def\m{\mathbf}

\DeclareMathOperator*{\argmax}{arg\,max}

\newcommand{\norm}[2]{\ensuremath{\left\|#1\right\|_{#2}}}

\newcommand {\bbmtx}{\begin{bmatrix}} %
\newcommand {\ebmtx}{\end{bmatrix}} %

\usepackage[normalem]{ulem}
\useunder{\uline}{\ul}{}

\usepackage[pagebackref=true,breaklinks=true,letterpaper=true,colorlinks,bookmarks=false]{hyperref}

\cvprfinalcopy %

\def\cvprPaperID{2864} %

\ifcvprfinal\fi

\begin{document}

\title{Pixel Adaptive Filtering Units}

\author[1]{Filippos Kokkinos\thanks{Work done during a summer internship at Huawei Noah's Ark Lab.}}

\author[2]{Ioannis Marras}
\author[2]{Matteo Maggioni}
\author[2]{Gregory Slabaugh}
\author[3]{Stefanos Zafeiriou}
\affil[1]{University College London, ${}^{2}$\text{Huawei Noah's Ark Lab}}
\affil[3]{Imperial College London}

\maketitle

\begin{abstract}
State-of-the-art methods for computer vision rely heavily on the translation equivariance and spatial sharing properties of convolutional layers without explicitly taking into consideration the input content. Modern techniques employ deep sophisticated architectures in order to circumvent this issue. In this work, we propose a Pixel Adaptive Filtering Unit (PAFU) which introduces a differentiable kernel selection mechanism paired with a discrete, learnable and decorrelated group of kernels to allow for content-based spatial adaptation. First, we demonstrate the applicability of the technique in applications where runtime is of importance. Next, we employ PAFU in deep neural networks as a replacement of standard convolutional layers to enhance the original architectures with spatially varying computations to achieve considerable performance improvements. Finally, diverse and extensive experimentation provides strong empirical evidence in favor of the proposed content-adaptive processing scheme across different image processing and high-level computer vision tasks. 
\end{abstract}

\section{Introduction}
Convolutions in general and convolutional layers in particular are fundamental operations underpinning  nearly any image processing and high level computer vision problem. The reasons behind their success, especially in deep learning, are two fold.  First, sharing of learnable kernels allows for a simple way to extract optimal features for any task. Second, these operations are implemented in an optimized and massively parallel fashion, allowing for fast processing of images and videos of arbitrary size. 

Once the kernels are learned, the processing of any input is, by design, identical. In detail, convolutions are content-agnostic since the same kernels are applied on all positions of any image irrespective of the content; a characteristic known as spatial sharing. However, in practice both the image content and the conditions vary considerably which inescapably forces the deployment of more training data, augmentations and additional learnable parameters, constituting a naive solution to the problem at hand. A paradigm-shifting approach is to use content-adaptive filtering which modulates the processing of an input according to statistical cues that are derived from an image or a dataset. As a result, different images will undergo a unique analysis based on the depicted content. 
\begin{figure}[t]
\begin{center}
  \includegraphics[width=1.1\columnwidth]{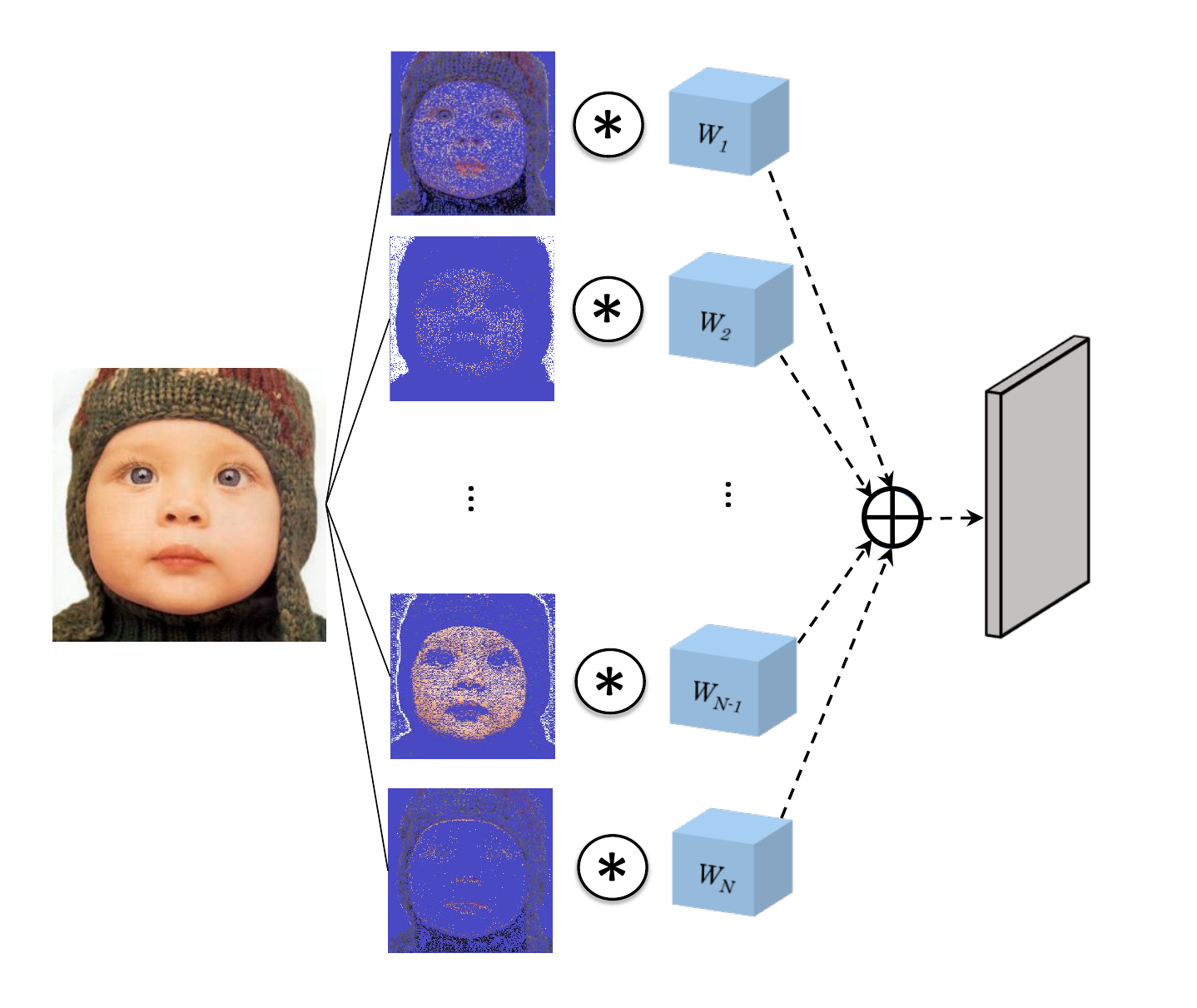}\vspace{-0.4cm}
  \caption{\textbf{Pixel Adaptive Filtering Unit (PAFU).} PAFU calculates as a first step a pixel wise selection of the kernels $\m W_i$. Afterwards, the kernels $\m W_i$ are applied on the original input according to the predicted selection pattern. Dark blue for illustration purposes indicates the area that the respective kernel will not be used. The output of the unit is a tensor with an arbitrary number of output channels. \vspace{-0.6cm}}
  \label{fig:pafu}
\end{center}
\end{figure}

In order to achieve content-adaptivity numerous methods have been proposed throughout the years. Non-Local Means~\cite{buades2005non} and bilateral filtering~\cite{tomasi1998bilateral} where among the first methods that utilized features and neighborhood statistics in order to modulate the filtering of an image according to the content. With the advent of deep learning, many of these spatially-varying techniques were adopted in a differentiable manner and used as layers in existing neural networks~\cite{Fanello_2014, Chang_2015, harley2017segmentation, Su_2019_CVPR}. Subsequently, a more recent trend is the development of deep neural networks that predict per pixel the convolutional kernels that need to be used~\cite{bako2017kernel, jia2016dynamic}, which in practice can be restricting since a large number of parameters must be learned in order to predict large convolutional kernels like those commonly found in an image classification deep CNN~\cite{he.delving}.  

In this work, we propose a new pixel adaptive filtering unit (PAFU), which performs differentiable selection of kernels from a discrete learnable and decorrelated group of kernels. As depicted in Fig.~\ref{fig:pafu}, the selection is made per pixel and thus the computational graph changes spatially according to the content of the input. The selection of the filters is done using a compact CNN network which is trained implicitly to select kernels based on features it extracts from the input. The end result of PAFU is the spatially-varying application of the kernels on the image or tensor to be filtered. This method provides the very appealing advantage of spatial adaptivity; a crucial component that is absent from the standard convolutional units like those commonly found in CNNs. Simultaneously, the set of kernels is regularized during training to be decorrelated and thus constitute a set of unique and diverse operators. In other words, this regularized group of kernels is enforced to have high variability hence avoiding the naive solution of a group with almost similar operators.%

We demonstrate the applicability of the method on both low and high level computer vision problems. First, the method is examined on an explicit linear domain for the problems of demosaicking and super-resolution for digital devices with limited resources (e.g. smartphones) where the runtime performance is of importance. In both problems, the results surpass the performance of competing linear methods while simultaneously achieve competitive accuracy with popular non-linear approaches. Then, we investigate PAFU as a replacement of convolutional layers in established deep CNNs proposed in literature for the aforementioned problems as well as classification and segmentation tasks. In all cases, there is a substantial performance improvement across a battery of tasks, providing strong empirical evidence for the need of spatial adaptivity and the benefits of selective filtering.  

\begin{figure*}[t]
  \centering
  \subfloat[Input\label{fig:toy_input}]{\includegraphics[width=0.2\textwidth,height=0.2\textwidth]{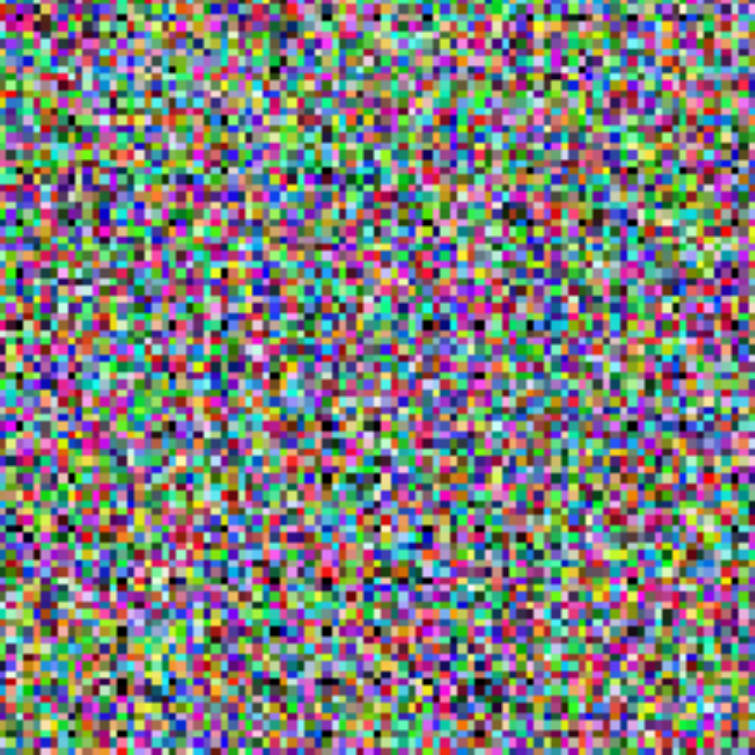}}
  \hspace{0.003cm}
  \subfloat[FCNN\label{fig:toy_fcnn}]{\includegraphics[width=0.2\textwidth,height=0.2\textwidth]{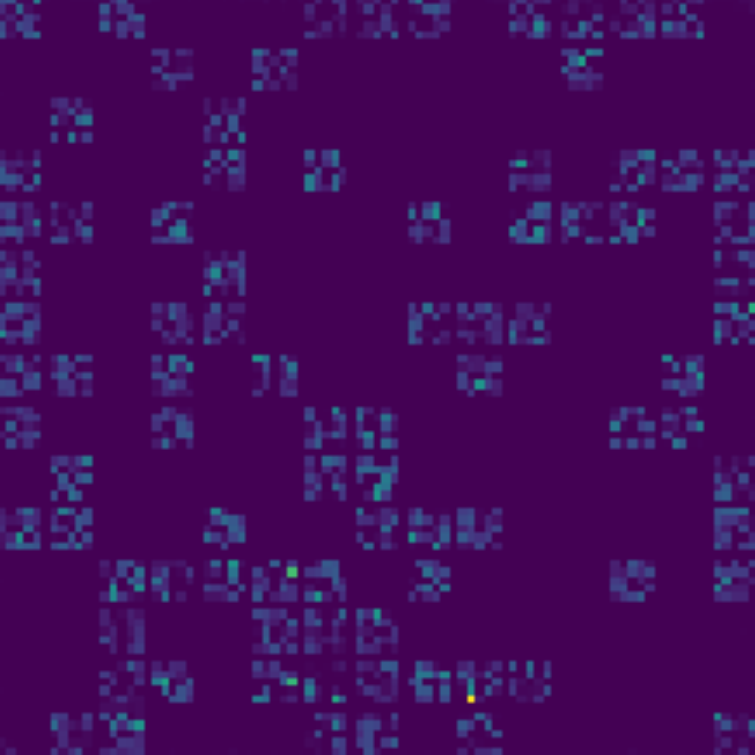}}
  \hspace{0.003cm}
  \subfloat[KPN\label{fig:toy_kpn}]{\includegraphics[width=0.2\textwidth,height=0.2\textwidth]{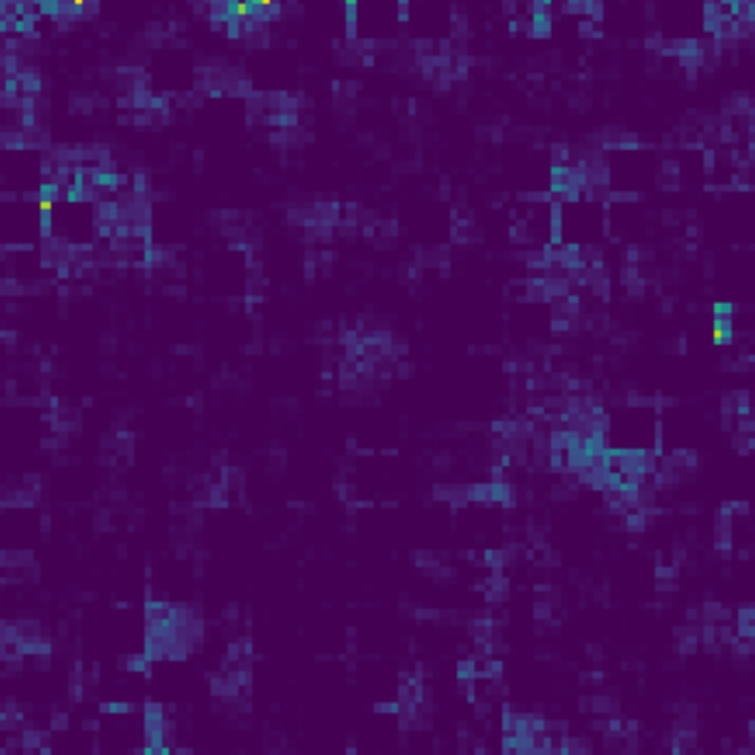}}
  \hspace{0.25cm}
  \subfloat[Filters Learned\label{fig:toy_filters}]{
  \subfloat{\includegraphics[width=0.099\textwidth]{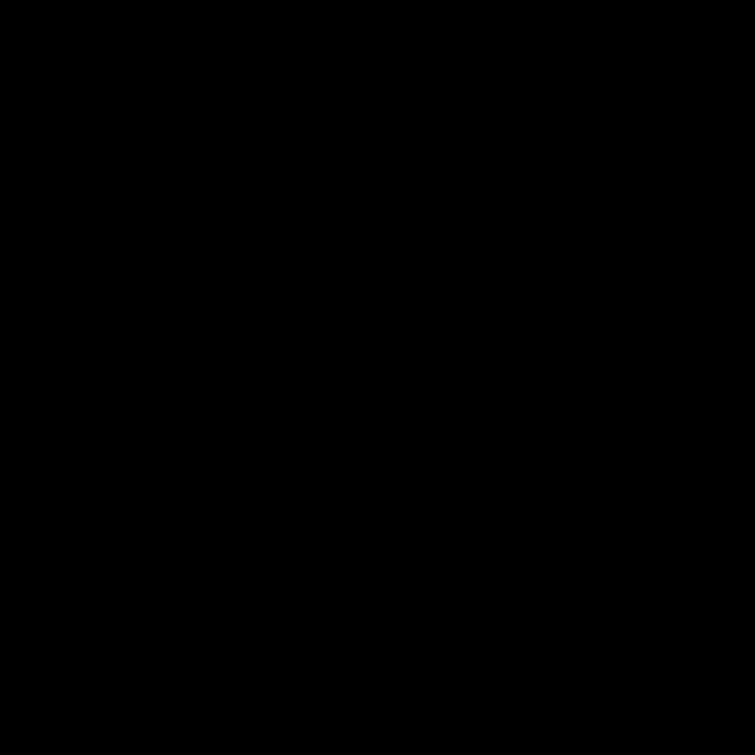}}
  \hspace{0.002cm}
  \subfloat{\includegraphics[width=0.099\textwidth]{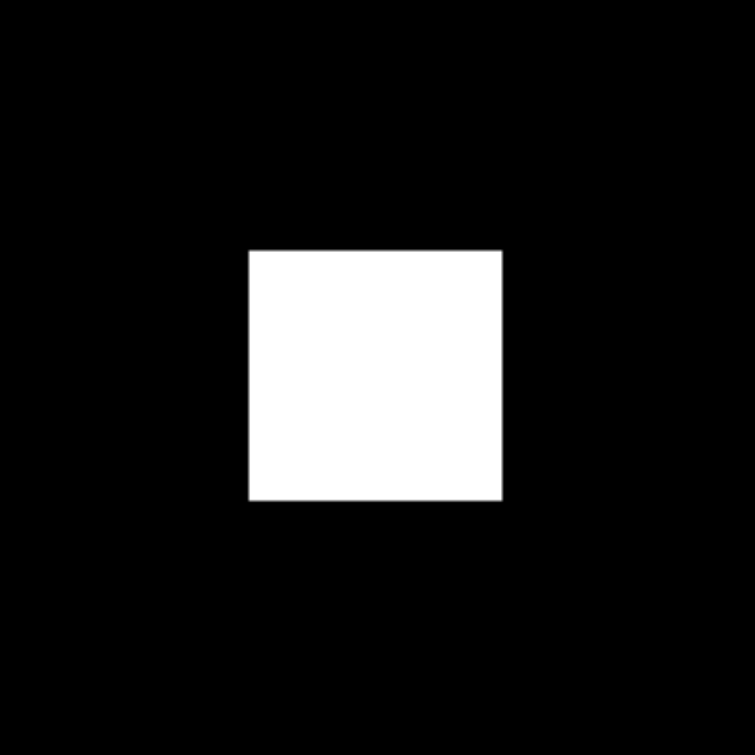}}
  }
  \\
  \vspace{-0.3cm}
  \subfloat[Ground-Truth\label{fig:toy_gt}]{\includegraphics[width=0.2\textwidth,height=0.2\textwidth]{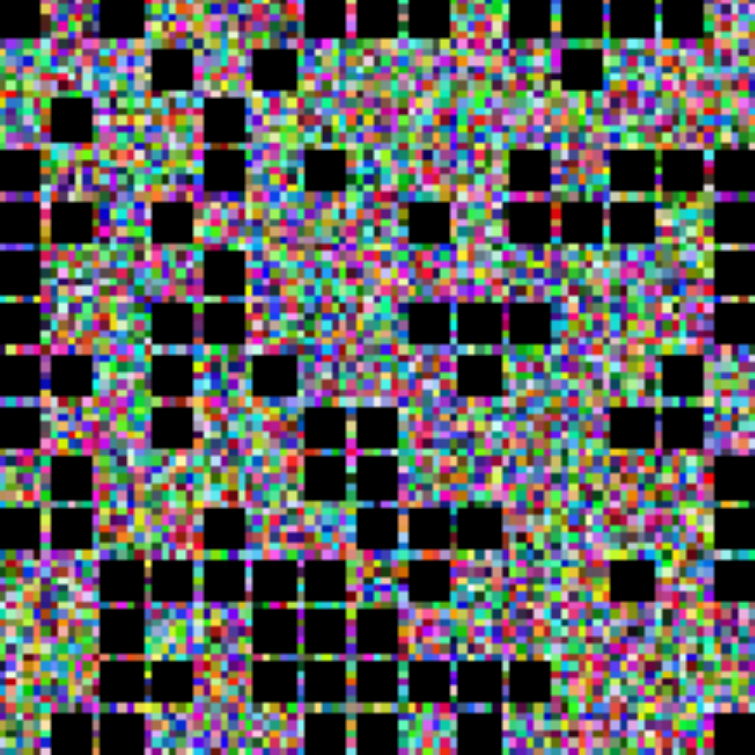}}
  \hspace{0.0035cm}
  \subfloat[Residual FCNN
  \label{fig:toy_fcnn_residual}]{\includegraphics[width=0.2\textwidth,height=0.2\textwidth]{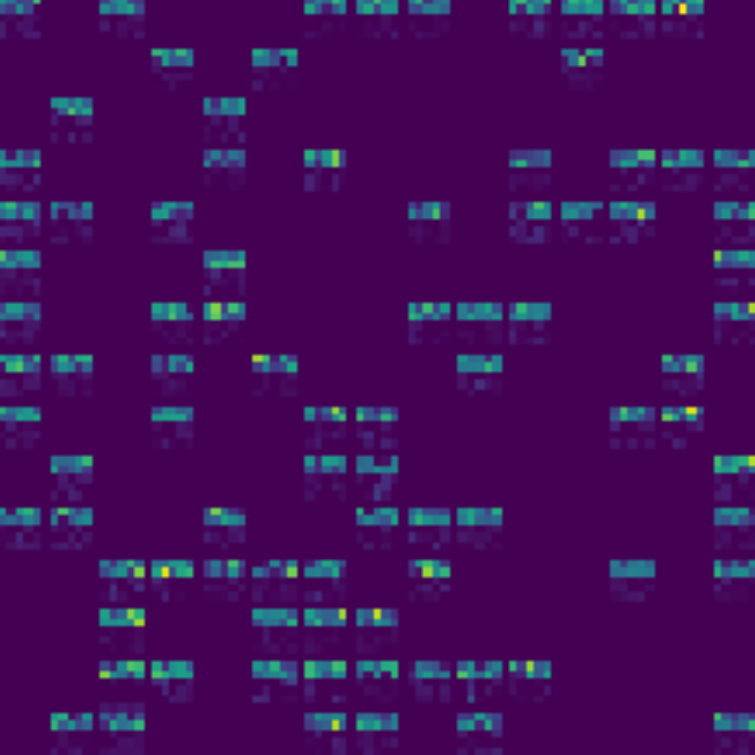}}
  \hspace{0.0035cm}
  \subfloat[PAFU (Ours)\label{fig:toy_pafu}]{\includegraphics[width=0.2\textwidth,height=0.2\textwidth]{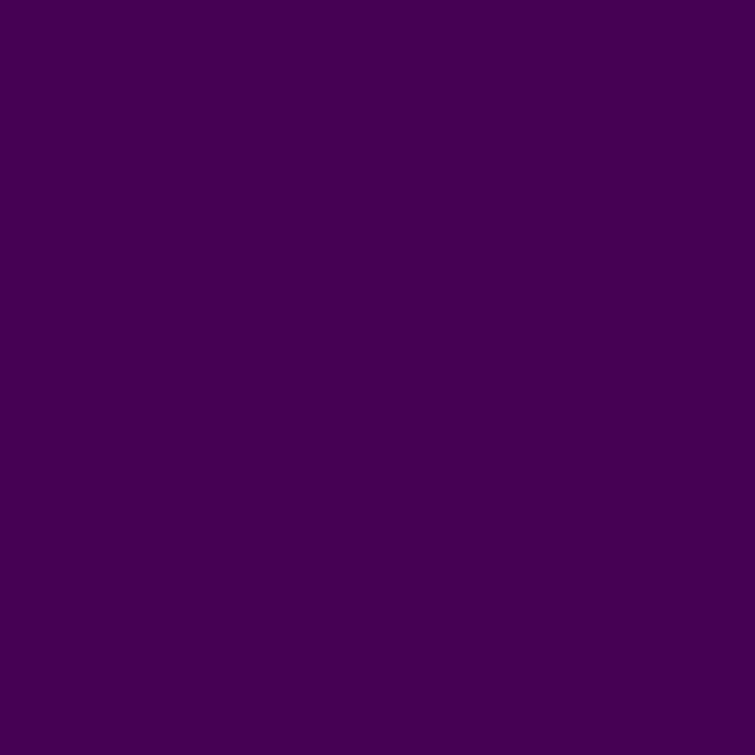}}
  \hspace{0.3cm}
  \subfloat[Filter Selection\label{fig:toy_fs}]{\includegraphics[width=0.2\textwidth,height=0.2\textwidth]{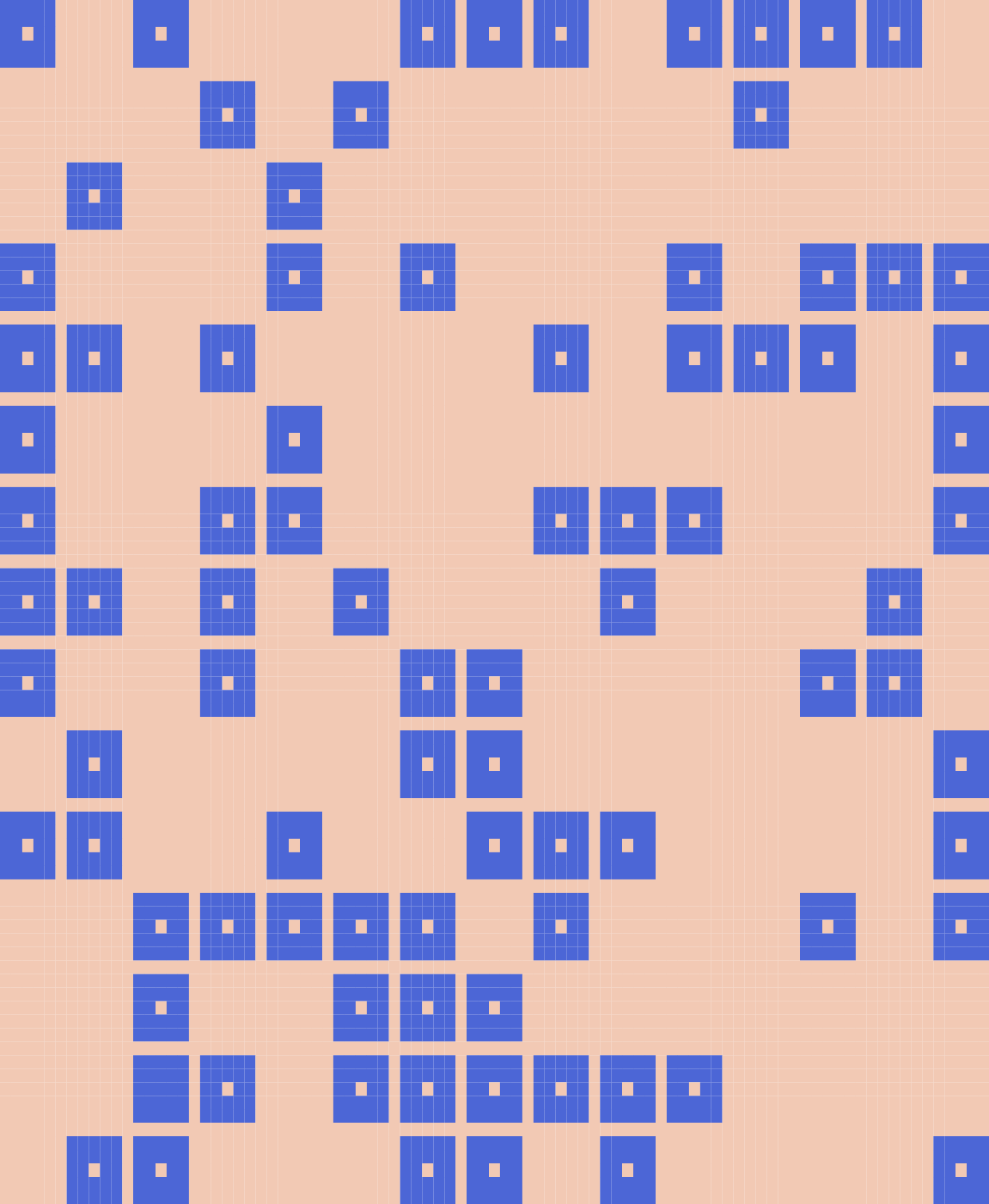}}
  \\
  \vspace{-0.1cm}
  \caption{In the SAD dataset, methods are trained to dilate all black pixels of the input image to 5x5 squares, while leaving every other pixel values unaltered. From the error plots (in b, c, h, and i), it is clear that all methods except ours struggle to learn the correct mapping. In the same time, we plot the two learned  utilized filters by the proposed method alongside the filter selection heatmap (in f and j respectively). \vspace{-0.2cm}}
  \label{fig:toy_dataset}
\end{figure*}

\section{Related Work}

\textbf{Spatially adaptive filtering:} Important research works on spatially adaptive filtering originate from image processing tasks and focus on modulating a filtering kernel in order to achieve visually pleasing reconstruction for tasks like image denoising and super-resolution. One of the earliest methods include non-local filtering~\cite{buades2005non,dabov2007image, Maggioni_2012}, guided image filtering~\cite{he_guided_2013}, bilateral filtering~\cite{tomasi1998bilateral} and propagation filters~\cite{Chang_2015}. A similar line of work aims to process different regions of an image with distinct kernels instead of modulating a unique one, which allows the construction of a task-specific over-redundant group of operators. Methods like RAISR~\cite{romano2016raisr}, BLADE~\cite{getreuer2018blade} as well as~\cite{stephanakis_1999, Fanello_2014} learn a discrete set of filters and select per pixel which filter should be applied using a selection mechanism based on local hand-crafted features.

Recently, many of these methods were adapted to be used alongside CNN layers and extend them by introducing spatially-varying modulation of computations. In~\cite{Harley_2017_ICCV, Su_2019_CVPR}, the authors adjust the convolutional kernels using Gaussian distances of embeddings from a separate network branch and apply them to the problems of depth map super-resolution and segmentation. Moreover, the embeddings used for kernel modulation are fine-tuned to minimize a metric learning loss along-side task specific losses. Jampani~\etal\cite{jampani_cvpr_2016} make heavy use of fast bilateral filtering techniques~\cite{sylvain_2006} in order to develop a learnable permutohedral lattice convolution module that was later tested on several tasks with great success. Most recently, a differentiable version of Non Local Means was introduced by Pl\"otz and Roth~\cite{Ploetz:2018:NNN} which selects the top-k closest neighbors of a centered pixel in a neighborhood as a part of a neural network and it was applied on the task of image denoising. 

\textbf{Kernel Prediction Networks (KPN):} A collection of recent works focuses on deep networks that predict the filtering kernels based on the content of the image~\cite{jia2016dynamic, bako2017kernel, Xue_2016}. All these works adjust the computation graph of the method to the input with high versatility, however, they require unavoidably a large amount of parameters. For example, a common convolutional layer in a standard classification network will require the prediction of at least half a million parameters\footnote{For example, the number of parameters for a kernel with size $3\times3\times256\times256$.} per pixel. Therefore, the computational burden becomes significant even for a single convolutional layer which limits the applicability of the kernel prediction networks to mainly linear scenarios~\cite{Mildenhall_2018_CVPR}. Moreover, according to~\cite{Mildenhall_2018_CVPR} kernel prediction networks require several tricks during training in order to ensure convergence which limits their applicability in practice.

\section{Pixel Adaptive Filtering Unit}
We propose Pixel Adaptive Filtering Units, which instead of modulating or predicting kernels, learn a group of kernels and a differentiable kernel selection mechanism based on a compact CNN. The formulation allows for fast and robust kernel selection with minimal overhead that depends mainly on the number of kernels of a group. 

\subsection{Spatially Invariant Convolution}
 The common way a convolutional layer is implemented in CNNs is a matrix vector product between a kernel $\mathcal{W} \in \mathbb{R}^{k \times k \times c_{out} \times c_{i n}}$ with support size $k$ and an input $\m x \in \mathbb{R}^{h \times w \times c_{in}}$. The kernel $\mathcal{W}$ transforms linearly the $c_{in}$ input channels to $c_{out}$ which by design takes into consideration the inter-channel correlations which are of importance for all modern CNN architectures. The output $\m y_{i,j} \in \mathbb{R}^{c_{out}}$ is formed as

\begin{equation}
\m y_{i,j} = 
  \sum_{\mathclap{a, b \in \mathcal{N}_k(i,j)}} 
        \mathcal{W}_{i-a, j-b} \; \m x_{a,b} 
\label{eq:standard_conv}
\end{equation}
where the neighborhood is defined as the set $\mathcal{N}_{k}(i, j)=\{a, b \bigm| | a-i|\leq k / 2,| b-j | \leq k / 2\}$. As it can be seen in Eq.~\eqref{eq:standard_conv} the same weights are applied on every position of $\m x$, a known property of convolutional layers which is known as spatial sharing. While this property has driven a lot of progress lately in computer vision tasks, as we will show, weight sharing across all positions is not effective to properly produce a dense spatially varying output. The intrinsic failure surfaces from the fact that the loss gradients from all image positions are fed into global kernels which are trained to minimize the error in all locations. The same problem arises in practice to a wide variety of problems which require a dense prediction or regression such as image segmentation, restoration and enhancement.

\subsection{Spatially Adaptive Dataset (SAD)}
In order to highlight the need for spatially adaptive processing of an image, we introduce a simple toy dataset, which we dub as Spatially Adaptive Dataset (SAD). The dataset is specifically designed to challenge spatially sharing methods like Fully Convolutional Neural Networks (FCNN) which severely under-perform on the task since they lack the appealing property of spatial adaptivity.

The dataset contains images of dimension $89\times89$ with random uniform color noise sampled from $(0,1]$ and black pixels sampled randomly from a predefined square grid with dimension $87\times87$ as in Fig.~\ref{fig:toy_input}. The aim is to dilate all black pixels to 5x5 black squares, which by design have no overlap as it is depicted in Fig.~\ref{fig:toy_gt}, while keeping all other values unaltered. The development of the SAD dataset allows the existence of an optimal minimal solution for the problem at hand, consisting of two filters; a Dirac and a zero filter. For pixels with random noise only the Dirac filter will leave them unaltered after filtering while the zero filter will dilate the black pixels to squares. We stress that any other solution cannot be characterized as both minimal and optimal. Note that black pixels located in the input image can be filtered with both filters without any difference as shown in Fig.~\ref{fig:toy_fs}.

The examined methods consist of a FCNN, a Residual FCNN, a KPN and our proposed PAFU. All of the aforementioned methods are trained to minimize the $\ell_1$ loss between the output and the groundtruth and have approximately the same number of parameters (nearly 35K). As it can be seen from Fig.~\ref{fig:toy_dataset}, methods that solely rely on spatially sharing convolutions can not yield adequate results and the reason lies in the fact that they apply the same kernels to every location of the input image. 

\subsection{Spatially Varying Convolution}
In PAFU, instead of applying the same kernel on all pixels, we selectively break the spatial sharing nature of convolutions by picking which kernels from a discrete group should be deployed on which locations of an image. The group of kernels  $\mathcal {\hat W} \in \mathbb{R}^{{n} \times k \times k \times c_{out} \times c_{in}}$ contains $n$ discrete kernels. In this case the spatially varying convolutional layer is defined as, 
\begin{equation}
\m y_{i,j} = 
  \sum_{\mathclap{a, b \in \mathcal{N}_k(i,j)}} 
     \mathcal {\hat W}_{\m z_{i,j},i-a, j-b} \; \m x_{a, b} 
\label{eq:pafu}
\end{equation}
where $\m z \in \mathbb{R}^{h \times w \times n}$ is an one-hot encoded index that indicates which kernel out of $n$ should be selected per pixel. The selection indices $\m z $ are predicted from a kernel selection mechanism $\phi$ given as input the image to be filtered, i.e. $\m z = \phi(\m x)$.  It is clear from Eq.~\ref{eq:pafu} that different regions of an image are filtered with distinct kernels from $ \mathcal {\hat W}$, thus selectively breaking the spatial sharing nature of convolutional layers by leveraging content lying in the input. 

Considering the illustrative example in Fig.~\ref{fig:toy_dataset}, a spatially-varying convolution is essential for the PAFU method to produce an output with zero error and learn the optimal filter set. The filter selection heatmap depicts which filters were selected per pixel and as it can be seen the selection between the two optimal filters is the proper one to produce minimal error. When compared to KPN, our method is capable to achieve lower error while KPN fails to predict a Dirac for the majority of the pixels that require this kernel.

With regards to computation overheard, the spatially varying convolution can be implemented in a parallel way using the standard \textit{im2col} and \textit{col2im} operations that break the spatial resolution of images into appropriate patches according to the kernel support sizes. Afterwards the patches can be easily filtered using a matrix vector operation per pixel as described in Eq.~\eqref{eq:pafu}. The same implementation is also a long known fast solution for the spatially invariant convolution, however modern computation libraries apply as set of low-level optimization techniques~\cite{Lavin_2016} to gain considerable execution time reduction. We note that there is a current lack of optimized implementations for spatially varying convolutional layers for the various hardware accelerators. However, their development is of high importance since it will allow for fast development and deployment of spatially adaptive techniques.

\subsection{Kernel Selection Mechanism}
An essential part of PAFU is the discrete selection of kernels based on the input content. Instead of performing the kernel selection based on hand-crafted rules as in~\cite{romano2016raisr}, we seek to learn the selection mechanism using available training data. In order to extract features of interest, we make use of a compact CNN that gets as input an image or embeddings $\m x \in \mathbb{R}^{h \times w \times c_{in}}$ and gives as output coefficients $\m f \in \mathbb{R}^{h \times w \times n}$ in order later to be used during the discrete kernel selection step.  

Given the coefficients $\m f$, we would like a discrete selection method which deploys on each pixel the most confident kernel $\m z_{i,j} = \argmax \m f_{i,j}$ from the group. However, the $\argmax$ function is non-differentiable and therefore not suitable to be used for this purpose. In this work, we use the Gumbel-Softmax method~\cite{jang2016categorical, MaddisonMT17}, which addresses the issue of discrete selection using a differentiable relaxation of the Gumbel-Max Trick~\cite{gumbel1954statistical, Papandreou_2011}, in order to obtain the selection indices $\m z$.  In the supplementary material, we provide an extended description of the adaption of Gumbel-Softmax method for the per-pixel selection of kernels.  

In all our experiments, the straight-through version of the Gumbel-softmax estimator is used, which during the forward pass discretizes the selection to be binary while the backward pass is calculated based on the continuous selection probabilities $\m z$. Similar to~\cite{Veit2019}, we observe that the straight-through estimator allows for faster convergence and intuitive kernel selection maps, no matter the apparent inconsistency between the forward and backward pass which theoretically leads to biased gradient estimation. The whole selection mechanism is trained implicitly to select the most suitable kernel from the group of kernels, which is also simultaneously learned, by minimizing a task-specific loss.

\begin{figure*}[t!]
\captionsetup[subfigure]{labelformat=empty}
  \centering
  \subfloat{\includegraphics[width=0.19\textwidth]{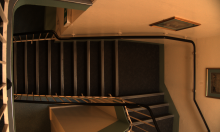}}
  \hspace{0.001cm}
  \subfloat{\includegraphics[width=0.19\textwidth]{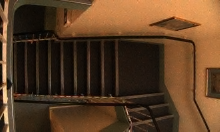}}
  \hspace{0.001cm}
  \subfloat{\includegraphics[width=0.19\textwidth]{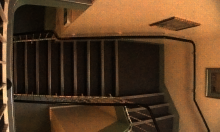}}
  \hspace{0.001cm}
  \subfloat{\includegraphics[width=0.19\textwidth]{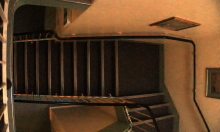}}
  \hspace{0.001cm}
  \subfloat{
    \hspace{-0.45cm}
    \begin{tabular}[b]{c}
        \vspace{-0.1cm}
        \includegraphics[width=.1\textwidth, height=0.97cm]{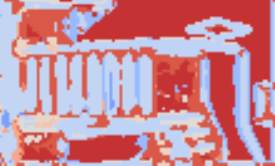} \\
        \vspace{-0.1cm}
        \includegraphics[width=.1\textwidth, height=0.97cm]{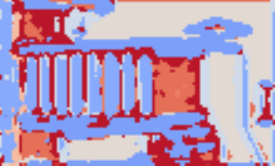}\\
    \end{tabular}
    \hspace{-0.65cm}
    \begin{tabular}[b]{c}
        \vspace{-0.1cm}
        \includegraphics[width=.1\textwidth, height=0.97cm]{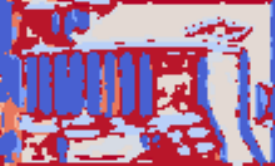} \\
        \vspace{-0.1cm}
        \includegraphics[width=.1\textwidth, height=0.97cm]{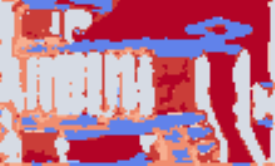}\\
    \end{tabular}
  }
  \\
  \vspace{-0.3cm}
  \subfloat{\includegraphics[width=0.19\textwidth]{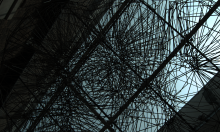}}
  \hspace{0.001cm}
  \subfloat{\includegraphics[width=0.19\textwidth]{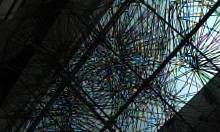}}
  \hspace{0.001cm}
  \subfloat{\includegraphics[width=0.19\textwidth]{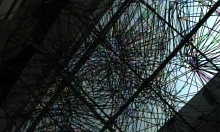}}
  \hspace{0.001cm}
  \subfloat{\includegraphics[width=0.19\textwidth]{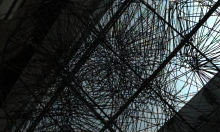}}
  \hspace{0.001cm}
  \subfloat{
    \hspace{-0.45cm}
    \begin{tabular}[b]{c}
        \vspace{-0.1cm}
        \includegraphics[width=.1\textwidth, height=0.97cm]{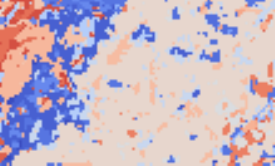} \\
        \vspace{-0.1cm}
        \includegraphics[width=.1\textwidth, height=0.97cm]{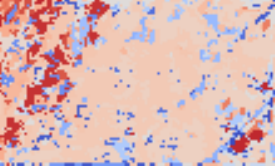}\\
    \end{tabular}
    \hspace{-0.65cm}
    \begin{tabular}[b]{c}
        \vspace{-0.1cm}
        \includegraphics[width=.1\textwidth, height=0.97cm]{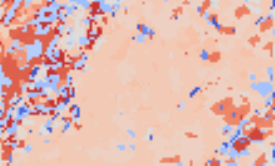} \\
        \vspace{-0.1cm}
        \includegraphics[width=.1\textwidth, height=0.97cm]{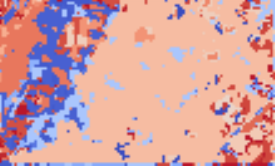}\\
    \end{tabular}
  }
  \\
  \vspace{-0.3cm}
  \setcounter{subfigure}{0}
  \subfloat[GT\vspace{-0.1cm}]{\includegraphics[width=0.19\textwidth]{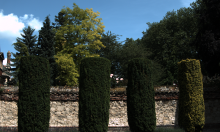}}
  \hspace{0.001cm}
  \subfloat[AHD~\cite{Hirakawa.2005}\vspace{-0.1cm}]{\includegraphics[width=0.19\textwidth]{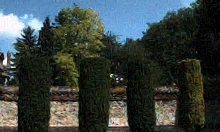}}
  \hspace{0.001cm}
  \subfloat[Khashabi~\etal~\cite{khasabi2014}\vspace{-0.1cm}]{\includegraphics[width=0.19\textwidth]{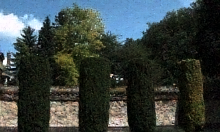}}
  \hspace{0.001cm}
  \subfloat[Ours\vspace{-0.1cm}]{\includegraphics[width=0.19\textwidth]{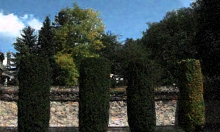}}
  \hspace{0.001cm}
  \subfloat[Selection\vspace{-0.2cm}]{
    \hspace{-0.45cm}
    \begin{tabular}[b]{c}
        \vspace{-0.1cm}
        \includegraphics[width=.1\textwidth, height=0.97cm]{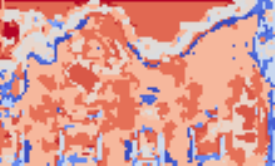} \\
        \vspace{-0.1cm}
        \includegraphics[width=.1\textwidth, height=0.97cm]{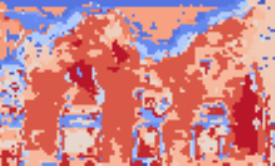}\\
    \end{tabular}

    \hspace{-0.65cm}
    \begin{tabular}[b]{c}
        \vspace{-0.1cm}
        \includegraphics[width=.1\textwidth, height=0.97cm]{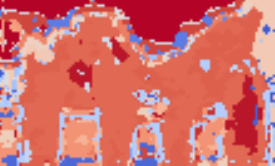} \\
        \vspace{-0.1cm}
        \includegraphics[width=.1\textwidth, height=0.97cm]{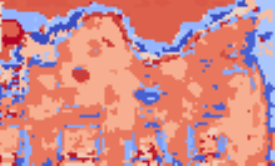}\\
    \end{tabular}
  }
  \\
 \vspace{-0.15cm}
\caption{\textbf{Linear Demosaicking.} Visual comparisons between three joint denoising and demosaicking methods. The kernel selection masks are broken into separate plots according to the Bayer Pattern (RGGB). It is clear that different kernels and proper selection mechanism is needed to successfully process each color of the Bayer. \vspace{-0.2cm}}\label{fig:linear_demosaicking}
\end{figure*}

\subsection{Enforcing Kernel Variability}
We found that a crucial component for robust selection is a regularizer enforcing variability to the group of learnable kernels. This form of regularization penalizes the naive solution where all kernels are identical and the kernel selection per pixel can be as good as chance. Simultaneously, kernels that are dissimilar with each other act as unique linear operators or feature extractors when used in deep neural networks. Their application will yield different results which increases the expressivity of the learnable group of kernels by suppressing any redundancies. In order to maximize dissimilarity and enforce variability we penalize the cosine distance between the kernels in a group. This is achieved by normalizing first and stacking afterwards all kernels in a matrix $\m W_f \in \mathcal{R} ^ {n\times n_p}$ where $n_p$ is the number of parameters in a kernel and then minimizing the regularization loss
\begin{equation}
    \label{eq:decorr_regularization}
    \ell_{\text {R}} =  \norm{\m W_f \m W_f^\top - \m I}{F}^2 .
    \vspace{-.1cm}
\end{equation}
In the case where we deploy kernels of different support sizes, the kernels are padded to the maximum support size before the formation of matrix  $\m W_f$. Every model presented is trained with decorrelation regularization alongside task-specific losses
\begin{equation}
\ell=\ell_{task}+\ell_{\text {R}}
\vspace{-.1cm}
\end{equation}
and in the event where more than one PAFU modules are utilized the regularization loss is the average of the individual losses.

\begin{table}[t]
\centering
\small
\begin{tabular}{lcc}
\specialrule{.1em}{.05em}{.05em} 
 & Noise-free & Noisy \\ \hline
Bilinear & 30.86 & 30.40 \\
D-MMSE~\cite{zhang2005color} & 38.82 & 36.67 \\
HA~\cite{hamilton1997adaptive} & 36.69 & 35.19 \\
AHD~\cite{Hirakawa.2005} & 37.23 & 35.77 \\
Khashabi~\etal~\cite{khasabi2014} & \textbf{39.39} & \textbf{37.80} \\ \hline
Ours: &  &  \\
\multicolumn{1}{c}{5x5} & 39.25 & 37.23 \\
\multicolumn{1}{c}{5x5\&7x7} & {\ul 39.38} & {\ul 37.37} \\
\specialrule{.1em}{.05em}{.05em} 
\end{tabular}%
\vspace{-0.1cm}
\caption{\textbf{Linear Demosaicking.} PSNR (in dB) results of both linear and non-linear methods for the noise-free and noisy scenario. Our method outperforms all linear variants and achieves competitive performance with a complex non-linear method without the need for any post-processing.\vspace{-0.2cm}}
\label{tbl:linear_demosaicking}
\end{table}

\section{Linear Case}
As a first step, the method is trained and tested as an one-shot linear solution for image processing tasks. Although  the kernel selection mechanism is non-linear, the application of the selected kernel per pixel constitutes a pure linear transformation. In detail, for each examined task the output is the result of the spatially-varying convolution of the per pixel selected kernel and the original input image, as described in Eq.~\eqref{eq:pafu}. Although the expressiveness of the method is restricted in the linear case, the end-result is an application which is fast and yet achieves competitive performance  with more complex and non-linear methods. 

Different PAFU modules are trained for the low-level image processing tasks of image super-resolution and demosaicking; the latter is further examined in a joint denoising and demosaicking formulation. A very appealing property of PAFU is that it allows for a group of kernels with different support sizes which is experimentally verified across all tasks. An extensive description of technical details can be found in the supplementary material.

\subsection{Demosaicking} \label{sec:linear_demosaicking}

We experiment with the MSR dataset~\cite{khasabi2014} which consists of 200 training, 100 validation and 200 testing images. The dataset is also examined with the presence of heteroskedastic Gaussian noise with unknown multiplicative and additive parameters. Following~\cite{kokkinos.2019} we augment the dataset with random flips and the method is trained to minimize the $\ell_1$ loss between output and ground-truth alongside our dissimilarity regularization. The optimization is performed in an alternating fashion where as a first step the filter selection unit is optimized and then the kernels; with a ratio one to three. As shown in Table~\ref{tbl:linear_demosaicking}, our approach yields impressive quantitative results in both variants without any domain specific techniques like directional interpolation and chromatic selection rules as in~\cite{zhang2011color, hamilton1997adaptive} or median iterative post-processing~\cite{Hirakawa.2005}. Moreover, while the method is restricted to be purely a linear application of kernels, the quantitative results are on par with~\cite{khasabi2014} which is based on a learnable cascade of Regression Tree Fields (RTFs) specifically designed for the image demosaicking task. Furthermore, PAFU is tested with different kernel support sizes and we experimentally find that multiple kernel sizes yield considerable performance improvement for the task without any additional computational overhead. As shown in Fig.~\ref{fig:linear_demosaicking}, PAFU method is capable to produce images with suppressed noise artifacts and reconstruct image without moire color distortions. 

\begin{figure*}[t]
\captionsetup[subfigure]{labelformat=empty}
\begin{centering}
\subfloat{\includegraphics[trim={0 0.5cm 0 1.8cm},clip, width=.25\textwidth]{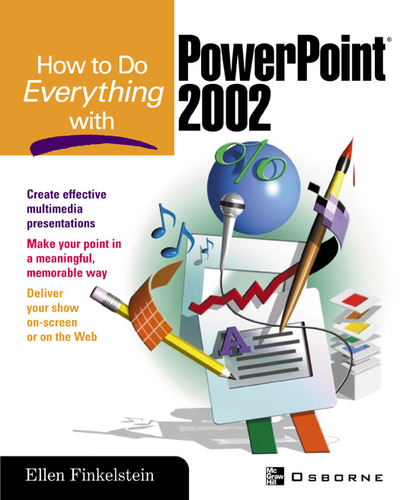}}~
\subfloat{\includegraphics[trim={0 0.5cm 0 1.8cm},clip,width=.25\textwidth]{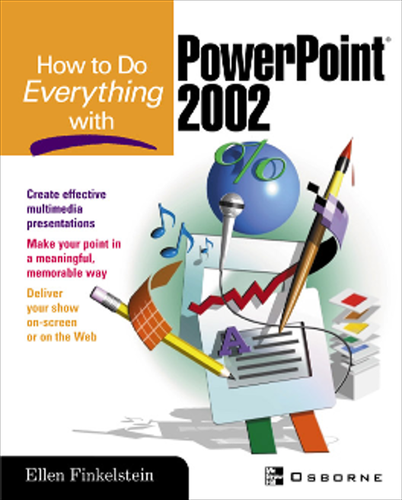}}~
\subfloat{\includegraphics[trim={0 0.5cm 0 1.8cm},clip,width=.25\textwidth]{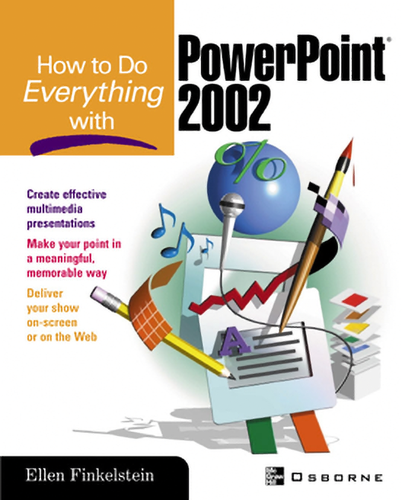}}~
\subfloat{\includegraphics[width=.25\textwidth, height=5cm]{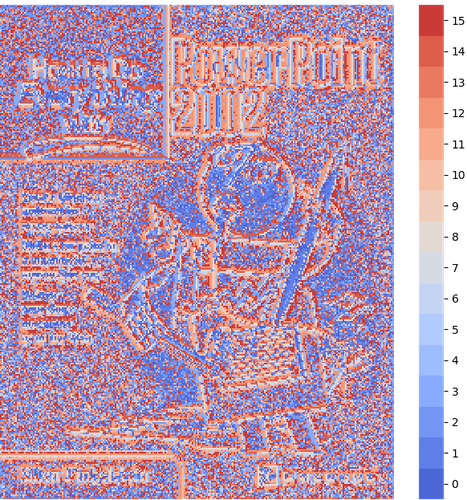}}~\\
\vspace{-2mm}
\setcounter{subfigure}{0}
\subfloat[\scriptsize GT]{\includegraphics[width=.25\textwidth]{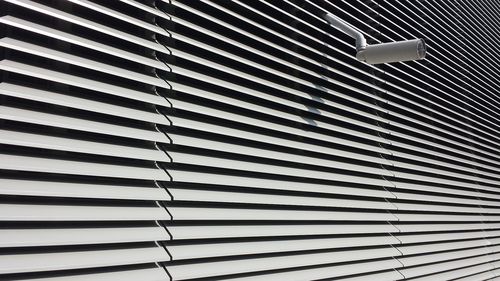}}~
\subfloat[\scriptsize Bicubic]{\includegraphics[width=.25\textwidth]{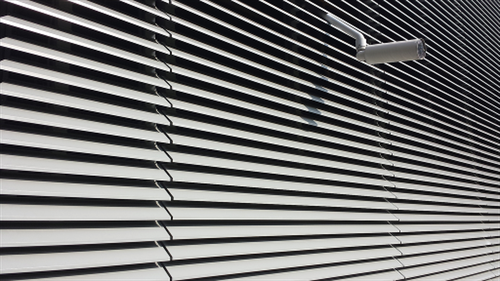}}~
\subfloat[\scriptsize Ours]{\includegraphics[width=.25\textwidth]{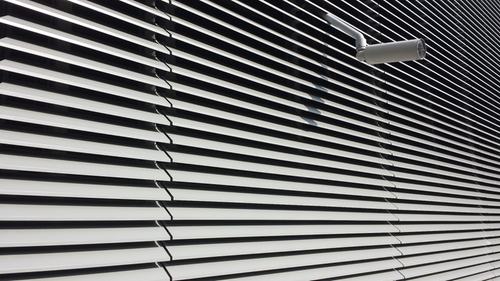}}~
\subfloat[\scriptsize Selection]{\includegraphics[width=.25\textwidth, height=2.5cm]{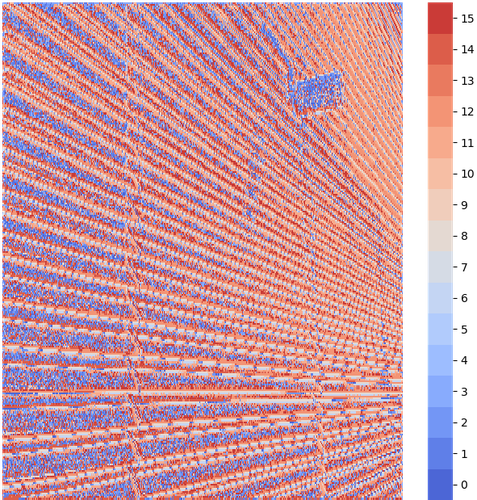}}~\\
\vspace{-0.25cm}
\caption{\textbf{Linear Super-Resolution.} Visual comparisons of super-resolution method alongside the kernel selection masks. The selection mechanism identifies the need to filter high-frequency content differently from smooth areas.\vspace{-0.2cm}}\label{fig:linear_sr}
\end{centering}
\end{figure*}

\begin{table}[t]
\centering
\resizebox{\textwidth}{!}{%
\begin{tabular}{llcccccc}
\specialrule{.1em}{.05em}{.05em} 
 & Supp. & \multicolumn{3}{c}{Set5} & \multicolumn{3}{c}{Set14} \\ \hline
 &  & x2 & x3 & x4 & x2 & x3 & x4 \\ \hline
Bicubic &  & 33.66 & 30.39 & 28.42 & 30.24 & 27.55 & 26.00 \\
RAISR~\cite{romano2016raisr} &  & 35.91 & 32.06 & 29.69 & 31.98 & 28.76 & 26.91 \\ \hline
\multirow{2}{*}{Ours} & 5x5 & 36.22 & 31.79 & 29.71 & {\ul 32.15} & 28.69 & 26.99 \\
 & 5x5\&7x7 & 35.70 & 31.90 & 29.74 & 31.81 & 28.75 & 27.01 \\ \hline
\multirow{2}{*}{Ours+} & 5x5 & \textbf{36.54} & {\ul 32.07} & {\ul 29.92} & \textbf{32.38} & {\ul 28.87} & {\ul 27.13} \\
 & 5x5\&7x7 & {\ul 36.08} & \textbf{32.12} & \textbf{29.96} & 32.08 & \textbf{28.89} & \textbf{27.15} \\
\specialrule{.1em}{.05em}{.05em} 
\end{tabular}%
}
\vspace{-0.1cm}
\caption{\textbf{Linear Super-Resolution.}The proposed method is capable to surpass in PSNR performance competing linear super-resolution methods. The "+" sign indicates a self-ensemble variant of our method. \vspace{-0.2cm}}
\label{tab:linear_sr}
\end{table}

\begin{table}[t]
\centering
\begin{tabular}{lcc}
\specialrule{.1em}{.05em}{.05em} 
Method & Noise-Free & Noisy \\ \hline
Bilinear & 30.86 & 30.40 \\
Khashabi~\etal~\cite{khasabi2014} & 39.39 & 37.80 \\
Ours Linear & 39.38 & 37.37 \\ \hline
Gharbi~\etal~\cite{Gharbi:2016:DJD:2980179.2982399} & 42.70 & 38.60 \\
Kokkinos~\cite{kokkinos.2019} & 42.77 & 40.08 \\ \hline
PAFU-Kokkinos & \textbf{42.78} & \textbf{40.15} \\
\specialrule{.1em}{.05em}{.05em} 
\end{tabular}
\vspace{-0.1cm}
\caption{\textbf{Demosaicking.} State-of-the art PSNR performance can be achieved by deploying PAFU in popular demosaicking methods. \vspace{-0.2cm}}
\label{tab:nonlinear_demosaicking}
\end{table}

\subsection{Super-Resolution} \label{sec:linear_sr}
It the same spirit with demosaicking, PAFU modules are examined as a linear solution for super-resolution similar to RAISR~\cite{romano2016raisr}, however we note that the most accurate methods for super-resolution are either deep or dictionary learning based~\cite{dong_2014_eccv, timofte2014}. Linear super-resolution models are trained using the DIV2K dataset~\cite{Agustsson_2017_CVPR_Workshops} which consists of 800 training and 100 validation images, following the training procedure used in~\cite{Lim_2017_CVPR_Workshops} in an alternating scheme as in Section.~\ref{sec:linear_demosaicking}. In order to speed up the method we perform both the kernel selection and the spatially variant convolution on the low resolution image. The final output consists of 12 channels in order to be super-resolved by the depth to space linear operation introduced in~\cite{shi2016real}. In this manner, PAFU is able to produce an image of higher resolution with minimal computation cost. Table~\ref{tab:linear_sr} showcases the performance improvement gained from a learned kernel selection when compared to the hand-crafted approach of RAISR~\cite{romano2016raisr}; an extensive qualitative comparison is provided in the supplementary material. We report both the results of our standard method for two different kernel support sizes as well as the standard self-ensemble variant as in ~\cite{Lim_2017_CVPR_Workshops}. Similarly to linear demosaicking, it is clear that various support sizes yield considerable performance improvement for scales larger than 2.

\begin{table}[t]
\centering
\small
\resizebox{0.9\textwidth}{!}{%
\begin{tabular}{lccccc}
\specialrule{.1em}{.05em}{.05em} 
 & Scale & Set5 & Set14 & B100 & Urban100 \\ \hline
\multirow{3}{*}{SRCNN~\cite{dong_2014_eccv}} & x2 & 36.66 & 32.45 & 31.36 & 29.5 \\
 & x3 & 32.75 & 29.30 & 28.41 & 26.24 \\
 & x4 & 30.48 & 27.50 & 26.90 & 24.52 \\ \hline
ESPCNN~\cite{shi2016real} & x3 & 33.00 & 29.42 & - & - \\ \hline
ESPCNN (ours) & x2 & 36.83 & 32.56 & 31.38 & 29.54 \\
 & x3 & 32.92 & 29.42 & 28.44 & 26.28 \\
 & x4 & 30.57 & 27.60 & 26.92 & 24.50 \\ \hline
 \multirow{3}{*}{PAFU-SRCNN} & x2 & 36.91 & 32.65 & 31.44 & 29.80 \\
 & x3 & 32.83 & 29.42 & 28.40 & 26.32 \\
 & x4 & 30.65 & 27.67 & 26.95 &24.63 \\ \hline
\multirow{3}{*}{PAFU-ESPCNN} & x2 & \textbf{37.10} & \textbf{32.76} & \textbf{31.57} & \textbf{30.04} \\
 & x3 & \textbf{33.08} & \textbf{29.53} & \textbf{28.53} & \textbf{26.48} \\
 & x4 & \textbf{30.87} & \textbf{27.77} & \textbf{27.01} & \textbf{24.73} \\
\specialrule{.1em}{.05em}{.05em} 
\end{tabular}%
}
\vspace{-0.1cm}
\caption{\textbf{Super-Resolution.} Quantitative PSNR results with a bicubic degradation model. The addition of a PAFU module in a super-resolution network allows for reconstruction of higher quality over the baseline network.\vspace{-0.2cm}}
\label{tab:deep_sr}
\end{table}

\section{Non-Linear Case}
Given the overall positive performance of our proposed method on the linear case, we investigate the module as a building block of deep neural networks to enable spatially-varying computations in CNN layers. The unit is used to replace several convolutional layers in standard neural networks for the tasks of image demosaicking, super-resolution, classification and segmentation. The diverse set of tasks allows us to examine the property of selectively breaking the spatially sharing nature of CNNs in dense prediction and regressions tasks as well as standard image classification.

\subsection{Demosaicking}
In order to allow spatially varying processing, we replace the input convolutional layer in the method of~\cite{kokkinos.2019, kokkinos.2018} with a PAFU unit which selects from a group of 16 kernels. Note that, the hyper-parameters are matched with the original network in order to keep the architecture identical. The model is trained from scratch following the same procedure with~\cite{kokkinos.2019}, alongside our dissimilarity regularization, using the MSR dataset for both noise-free and noisy scenarios. As shown in Table~\ref{tab:nonlinear_demosaicking} the PSNR improvement is substantial in the joint denoising and demosaicking problem and constitutes a new state-of-the art performance for the task. 

\subsection{Super-Resolution}
In Section~\ref{sec:linear_sr} a PAFU unit was trained to solve the problem of super-resolution in a linear manner, however it is a well-known fact that super-resolution tasks require non-linear treatment in order to achieve visually pleasing results. While there is an abundance of deep learning based super-resolution methods, we focus mainly on a narrow set of techniques that have less than half a million parameters and their PSNR performance across different test sets is not saturated, as it is common in all recent super-resolution works. We propose an extension of ESPCNN~\cite{shi2016real} and SRCNN~\cite{dong_2014_eccv} which use a PAFU unit in the output in order to add spatial adaptivity in the module responsible for the RGB reconstruction of the image. Our spatial adaptive unit deploys a group of 16 kernels and the selection is done on the embeddings generated before the final output. Originally, ESPCNN upsampled only the luminance channel of an image and was trained to minimize the $\ell_2$ loss between the super-resolved image and the groundtruth. We extend the original method to perform super-resolution of the RGB image by changing respectively the numbers of input and output channels while keeping the rest of the CNN architecture identical. In Table~\ref{tab:deep_sr} the proposed modification trained on the DIV2K dataset following the training procedure described in~\cite{Lim_2017_CVPR_Workshops} yields similar performance with the results reported in literature. As shown in Table~\ref{tab:deep_sr}, both networks enhanced with spatial adaptivity outperform significantly the original architectures. Extensive qualitative comparisons between the aforementioned methods can be located in the supplementary material.

\begin{table}[t]
\centering
\resizebox{0.95\textwidth}{!}{%
\begin{tabular}{llcccc}
 \specialrule{.1em}{.05em}{.05em} 
\multicolumn{1}{c}{} & \multicolumn{1}{c}{} & \multirow{2}{*}{Depth} & \multicolumn{2}{c}{ResNet} & \multirow{2}{*}{PAFU-ResNet} \\
\multicolumn{1}{c}{} & \multicolumn{1}{c}{} &  & Orig. & By~\cite{gao_2016} &  \\ \hline
C10 & \multicolumn{1}{c}{} & 110 & 6.61 & 6.41 & \textbf{5.33} \\
C100 & \multicolumn{1}{c}{} & 110 & - & 27.76 & \textbf{26.02} \\ \hline
\multirow{4}{*}{INet} & Top1 & 50 & 24.7 & - & \textbf{23.42} \\
 & Top5 & 50 & 7.8 & - & \textbf{6.82} \\ \cline{2-6} 
 & Top1 & 101 & 23.6 & - & \textbf{22.08} \\
 & Top5 & 101 & 7.1 & - & \textbf{5.9} \\
  \specialrule{.1em}{.05em}{.05em} 

\end{tabular}
}
\vspace{-0.1cm}
\caption{\textbf{Image Classification Results.} Test Error of ResNet based architectures on three different datasets. Spatially adaptivity increases significantly the classification quality across all datasets. \vspace{-0.3cm}}
\label{tab:image_classification}
\end{table}

\subsection{Image Classification}

\begin{figure*}[htbp!]
\captionsetup[subfigure]{labelformat=empty}
\begin{centering}
\subfloat{\includegraphics[width=.2\textwidth]{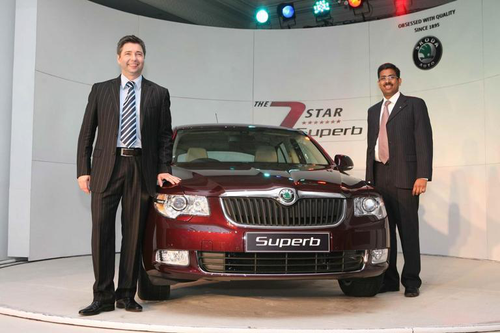}}~
\subfloat{\includegraphics[width=.2\textwidth]{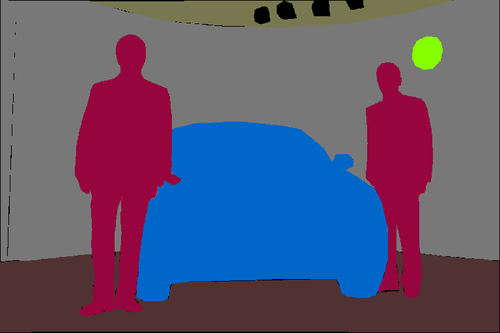}}~
\subfloat{\includegraphics[width=.2\textwidth]{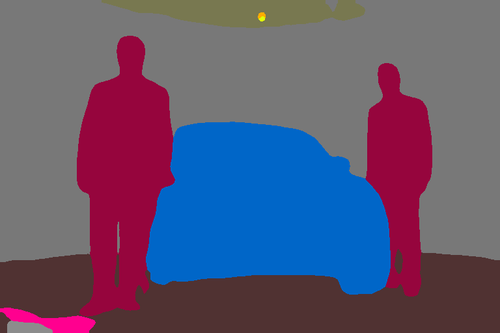}}~
\subfloat{\includegraphics[width=.2\textwidth]{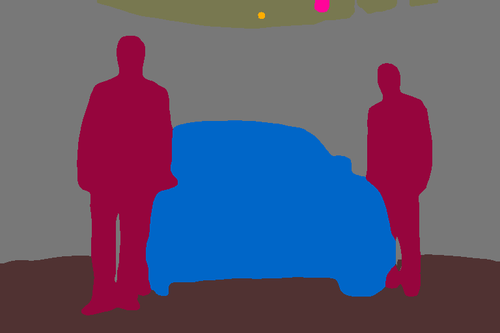}}~
\subfloat{\includegraphics[width=.2\textwidth, height=2.35cm]{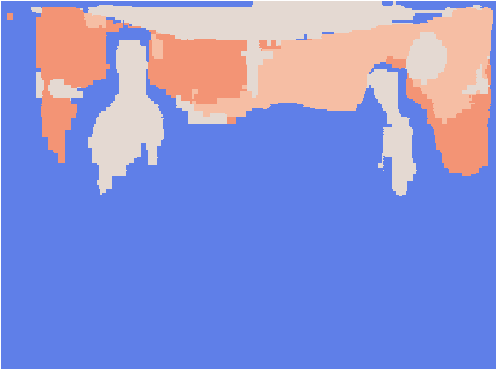}}~\\
\vspace{-0.35cm}
\subfloat{\includegraphics[width=.2\textwidth]{latex/images/segmentation/ADE_val_00000034_inp.png}}~
\subfloat{\includegraphics[width=.2\textwidth]{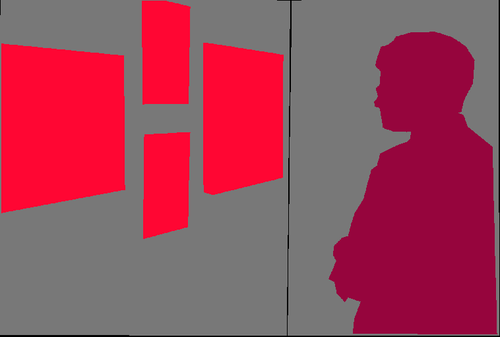}}~
\subfloat{\includegraphics[width=.2\textwidth]{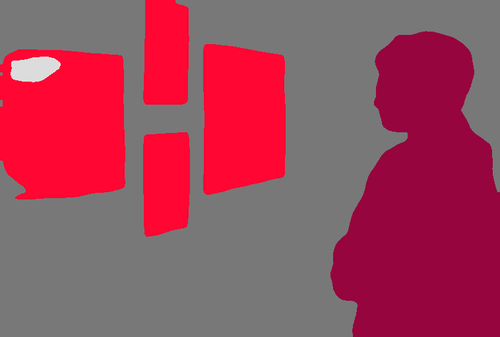}}~
\subfloat{\includegraphics[width=.2\textwidth]{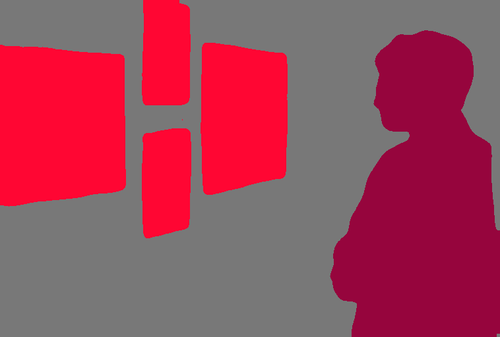}}~
\subfloat{\includegraphics[width=.2\textwidth, height=2.38cm]{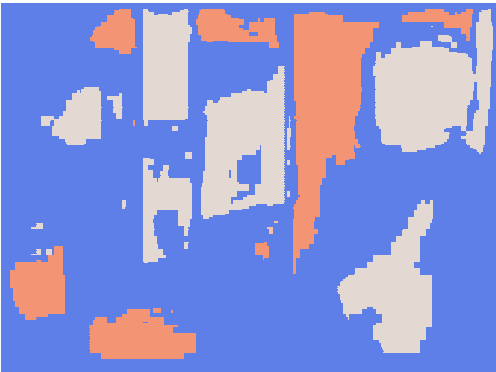}}~\\
\vspace{-0.35cm}
\setcounter{subfigure}{0}
\subfloat[\scriptsize Input]{\includegraphics[width=.2\textwidth]{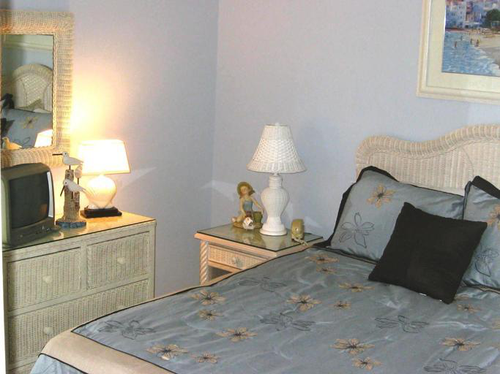}}~
\subfloat[\scriptsize GT]{\includegraphics[width=.2\textwidth]{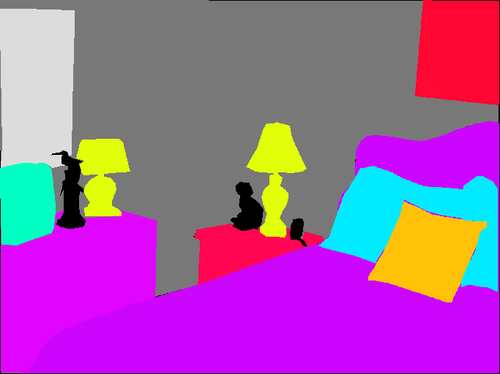}}~
\subfloat[\scriptsize PSPNet~\cite{zhao2017pyramid}]{\includegraphics[width=.2\textwidth]{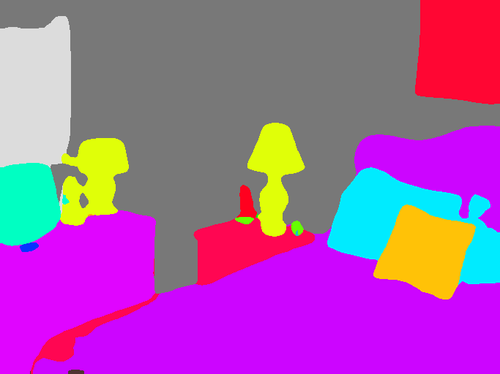}}~
\subfloat[\scriptsize Ours]{\includegraphics[width=.2\textwidth]{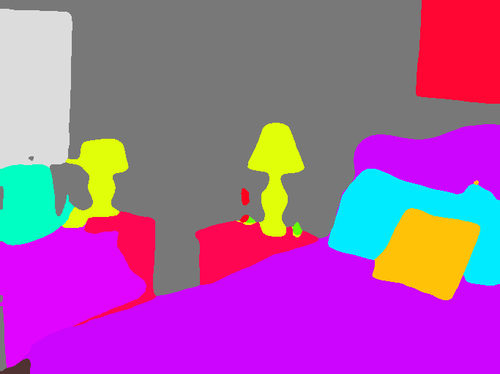}}~
\subfloat[\scriptsize Selection]{\includegraphics[width=.2\textwidth, height=2.64cm]{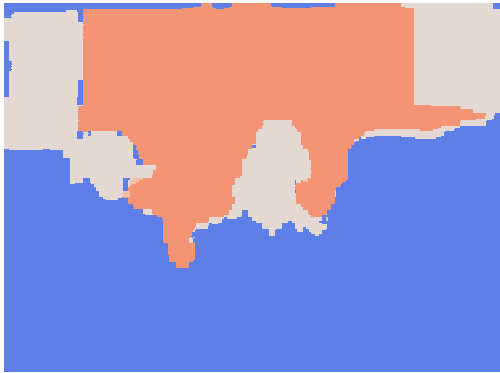}}~\\
\vspace{-0.15cm}
\caption{\textbf{Segmentation Results.} Examples of results from the validation set of ADE20K between the PPM and our proposed spatial adaptive extension of PSPNet. Next to our results with visualize the kernel selection masks where each color indicates different kernels.\vspace{-0.2cm}}\label{fig:seg}
\end{centering}
\end{figure*}

For every image classification experiment we used the standard ResNet architectures described in~\cite{he.2015} for the datasets CIFAR-10, CIFAR-100 and ImageNet~\cite{imagenet_cvpr09}. In every model we insert our module in the first convolutional unit in each of the three residual stacks~\cite{he.2015} and select kernels from a group with size 16. The results are presented in Table~\ref{tab:image_classification}. All models were trained according to the training procedure described in~\cite{gao_2016} for fair comparison. Note that our spatially adaptive filtering units improve the accuracy of image classification datasets significantly across  all datasets.

Empirically we find that PAFU modules in the first two stacks select consistently only one kernel from the group and only the third stack selects more than one kernels. This phenomenon can be explained from the fact that image classification tasks aim to develop a unique representation of an image based on extracted features in order to be classified later using a  dense MLP. Therefore, spatial sharing is a core component especially in early feature extraction stages where the features typically represent low level patterns like edges. However, in later stages some extracted high-level features might not help the proper classification of an image and thus our selective mechanism weighs accordingly their contribution. Note that, this is an advantage of our method, since during inference units that select only one kernel can be replaced with a single convolutional layer which filters the input with the selected kernel from the group.

\subsection{Semantic Segmentation}

\begin{table}[b]
\centering
\resizebox{\textwidth}{!}{%
\begin{tabular}{lcccc}
 \specialrule{.1em}{.05em}{.05em} 
 & Multi-Scale & Mean IoU(\%) & Pixel Acc.(\%) \\ \hline
\multirow{2}{*}{ResNet50+PPM} & No & 41.26 & 79.73 \\
 & Yes & 42.14 & 80.13 \\ \hline
\multirow{2}{*}{ResNet50+PAFU-PPM} & No & {\ul 41.89} & {\ul 80.17}\\
 & Yes & \textbf{42.49} & \textbf{80.60}\\
 \specialrule{.1em}{.05em}{.05em} 
\end{tabular}%
}
\caption{\textbf{ADE20k results.} Results on the validation set with and without  the PAFU unit. Spatial adaptivity allows the same network architecture to achieve higher accuracy in dense pixel prediction tasks.}
\label{tab:segmentation}
\end{table}

We test PAFU on the MIT Scene Parsing Benchmark Dataset~\cite{zhou2017scene} which contains 22k scene-centric images with 150 different semantic categories. The dataset is split into 20k images for training and 2k image for validation purposes. The network we use for baseline is the Pyramid Scene Parsing Network (PSPNet) proposed in~\cite{zhao2017pyramid} with a ResNet50 backbone feature extractor, which aggregates context through a four level pyramid module in order to create a global representation of the image. Our proposed spatially adaptive extension replaces the convolutional layer, that is used on the concatenated pyramid pooling global features, with a PAFU unit that selects among 8 kernels and train the whole architecture with the training procedure described in~\cite{zhao2017pyramid}. While the final embeddings already encapsulate information of importance from a wide range of scales and a diverse set of features, we empirically find that spatial adaptivity increases significantly both the pixel level accuracy as well as the mean Intersection of Union (IoU), as reported in Table~\ref{tab:segmentation}.  As shown in Fig.~\ref{fig:seg}, whole areas of the input tensor are filtered with distinct kernels that in return provide more accurate and sharper dense prediction over the baseline method. In the supplementary material we provide more qualitative illustrations alongside the predicted selection masks.

\section{Conclusions}
In this work, we propose a novel pixel adaptive filtering unit that can selectively break the spatial sharing nature of convolutional layers. We show that PAFU can be deployed both as a standalone unit or as a module in deep neural networks. We demonstrate how PAFU can be used  with great success for a wide collection of tasks ranging from low-level image processing to high-level computer vision tasks.

{\small
\bibliographystyle{ieee_fullname}
\bibliography{egbib}
}

\end{document}